\documentclass[conference]{IEEEtran}

\usepackage{amsmath, amssymb}
\usepackage{hyperref}
\usepackage{xcolor}
\usepackage{array}
\usepackage{graphicx}

\title{Multilabel Classification for Lung Disease Detection: Integrating Deep Learning and Natural Language Processing\\
{\footnotesize \textsuperscript{*}Note: All authors had equal contribution to this work.}}
\makeatletter
\newcommand{\linebreakand}{%
  \end{@IEEEauthorhalign}
  \hfill\mbox{}\par
  \mbox{}\hfill\begin{@IEEEauthorhalign}
}
\makeatother
\author{
    \IEEEauthorblockN{Maria Efimovich\textsuperscript{*}}
    \IEEEauthorblockA{
        Mechanical Engineering \\
        Staten Island Tech High School \\
        New York, US \\
        \texttt{mefimovich11@gmail.com}
    }
    \and
    \IEEEauthorblockN{Jayden Lim\textsuperscript{*}}
    \IEEEauthorblockA{
        Computer Science \\
        Monta Vista High School \\
        California, US \\
        \texttt{jaydenclim@gmail.com}
    }\and
    \IEEEauthorblockN{Vedant Mehta\textsuperscript{*}}
    \IEEEauthorblockA{
        Computer Science \\
        Lambert High School \\
        Georgia, US \\
        \texttt{veds.mehta@gmail.com}
    }\linebreakand
    
    \IEEEauthorblockN{Ethan Poon\textsuperscript{*}}
    \IEEEauthorblockA{
        Electrical Engineering \\
        Edison Academy Magnet School \\
        New Jersey, US \\
        \texttt{ewpoon007@gmail.com}
    }
}

\begin{document}
\maketitle

\begin{abstract}
Classifying chest radiographs is a time-consuming and challenging task, even for experienced radiologists. This provides an area for improvement due to the difficulty in precisely distinguishing between conditions such as pleural effusion, pneumothorax, and pneumonia. We propose a novel transfer learning model for multi-label lung disease classification, utilizing the CheXpert dataset with over 12,617 images of frontal radiographs being analyzed. By integrating RadGraph parsing for efficient annotation extraction, we enhance the model's ability to accurately classify multiple lung diseases from complex medical images. The proposed model achieved an F1 score of 0.69 and an AUROC of 0.86, demonstrating its potential for clinical applications. Also explored was the use of Natural Language Processing (NLP) to parse report metadata and address uncertainties in disease classification. By comparing uncertain reports with more certain cases, the NLP-enhanced model improves its ability to conclusively classify conditions. This research highlights the connection between deep learning and NLP, underscoring their potential to enhance radiological diagnostics and aid in the efficient analysis of chest radiographs.
\end{abstract}


\section{Introduction}
Coupling healthcare with artificial intelligence (AI) automation has recently gained much traction due to its potential to greatly improve patient outcomes, reduce costs in the medical industry, and streamline efficiency. The leading application of AI in healthcare is in diagnostics and medical imaging, where AI algorithms are programmed to analyze medical scans, such as CTs, X-rays, and MRIs \cite{Ramalingam2023}. With the ever-growing access to scientific and medical data volumes, the intersection of medicine and AI has ignited new enthusiasm and holds great promise for significant advancements \cite{Rattan2022}. Burnout is characterized by a variety of symptoms, including a loss of enthusiasm for work, high emotional exhaustion, and high depersonalization on the symptoms side, as well as the more alarming workplace-oriented symptoms such as loss in productivity, high medical turnover, early retirement, and increasing health care costs \cite{Chetlen2018}. Generally, physicians have a very high reported burnout rate (around 45\% as of 2024), and radiology departments follow a similar trend \cite{AMA2024}. Alleviating radiologists’ demanding workload may offer a potential solution to mitigate this burnout, reduce errors, and remove human error within medicine. \par
Chest X-rays (CXRs) (with more than 2 billion scans captured worldwide every year) have been used as the baseline chest imaging scan for over a century due to their widespread availability, short scan time, cost-effectiveness, and low radiation exposure \cite{Ahmad2023}. CXRs can also diagnose a myriad of chest disorders, ranging from common lung infections such as pneumonia, tuberculosis, and lung cancer to cardiomegaly, a condition in which the heart is enlarged \cite{MayoClinic} There have been growing efforts to create AI systems to assist radiologists in analyzing CXRs. Machine learning, a subarea of the broader field of AI, involves learning patterns from fed data to enable accurate predictions and classifications from the model. The majority of the applications of these deep learning models have been limited in scope, though, as the potential risks behind AI bias may lead to inaccuracy \cite{Ahmad2023}. 
Classifying CXRs is time-consuming and challenging, even for radiologists to precisely distinguish between pleural effusion, pneumothorax, and pneumonia. In this paper, we propose a transfer learning model for multi-label lung disease classification, where Radgraph parsing is used to parse for annotations within the CheXpert dataset’s frontal radiographs. 
There have been growing efforts to create AI systems to assist radiologists in analyzing CXRs. Machine learning (ML), a subarea of the broader field of AI, involves learning patterns from fed data to enable accurate predictions and classifications from the model \cite{Ahmad2023}. The majority of the applications of these models have been limited in scope, though, as the potential risks behind AI bias may lead to inaccuracy \cite{Ahmad2023}. \par
Deep learning (DL), a specialized subdomain of ML, has been one of the more prominent research trends due to its notable success with classification - it utilizes transformations and graph technologies in order to build complex, multi-layer models. Convolutional neural networks (CNNs) are the most commonly used and popular of the DL networks, as they’re able to automate the featuring learning process for tasks, allowing for simultaneous learning and classification in a single step, unlike ML. With that, DL models have quickly become a widespread approach for situations like image recognition, topic classification, and natural language understanding. \par
CNNs contain three distinct sets of layers: convolutional, pooling, and fully-connected. For image classification, the first input layer contains the raw pixel values of the image tested. The convolutional layers, defined by their corresponding kernel sizes (the filter dimension size), stride lengths (the step size of the kernels as it moves from the input to calculating product points and finding output pixels), and padding (the thickness of the 0-th frame around the input feature map), then compute the output of the neurons by connecting them to the local, original regions of the input and calculating the scalar products between their weights and the corresponding input region.\par
\begin{figure}[h]
    \centering
    \includegraphics[width=0.8\linewidth]{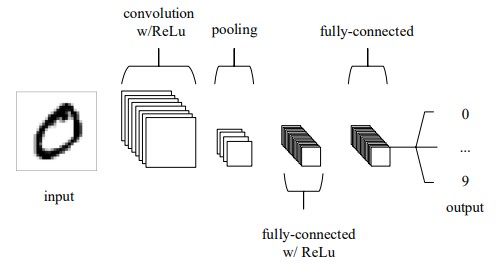}
    \caption{Convolutional neural network (CNN)}
    \label{fig:cnn}
\end{figure}
The rectified linear unit, abbreviated as the ReLU, applies an activation function from the previous output. Other commonly used activation functions include Sigmoid, Tanh, Leaky ReLU, and Parametric Linear Units (Purwono et al., 2023). The pooling layer reduces the number of parameters and creates down-sampled representations, thus lessening the overall computational cost and load. The third layer is the fully connected layer, typically the last layer in a CNN, that receives the input from the final pooling layer (Yamashita et al., 2018). \par
\begin{figure}[h]
    \centering
    \includegraphics[width=0.8\linewidth]{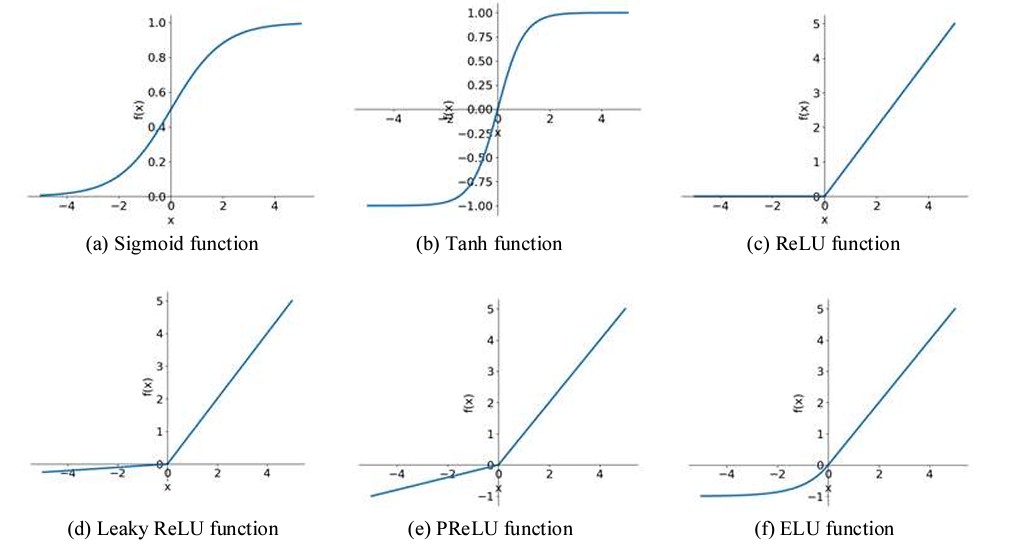}
    \caption{6 activation functions: Sigmoid, Tanh, ReLU, Leaky ReLU, PReLU, and ELU}
    \label{fig:cnn}
\end{figure}
Previously limited by the rigidity of rule-based algorithms, NLP has been transformed by integrating AI and deep learning, offering significantly enhanced potential and versatility in handling complex tasks (Chen et al., 2024). NLP tools have allowed for efficient information extraction from various unstructured text documents, such as radiology reports and clinic letters in the healthcare field. The future shows great promise in implementing algorithms that utilize automated radiology reporting. With the workload of diagnostic radiologists rapidly growing, NLP could greatly reduce the burden radiologists face with analyzing an exceeding number of scans (Aramaki et al., 2022). Various information extraction schemas have already been created for NLP-targeting in radiographs (MIMIC-CXR and CheXpert), but the labels they create do not capture the critical, fine-grained information contained in a radiology report, like specific entities and the relations between them. Radgraph, proposed by Jain et. al, addresses these limitations by including dense annotations for these two objects, paving the path for more comprehensive integration of radiology reports into multi-modal AI models. \par
\begin{figure}[h]
    \centering
    \includegraphics[width=0.8\linewidth]{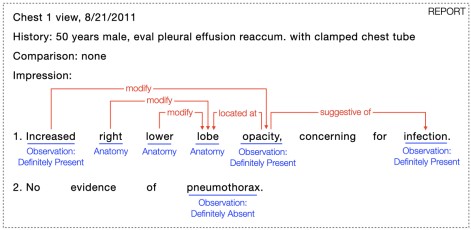}
    \caption{Radgraph annotations from a report}
    \label{fig:radg}
\end{figure}
\begin{figure}[h]
    \centering
    \includegraphics[width=0.6\linewidth]{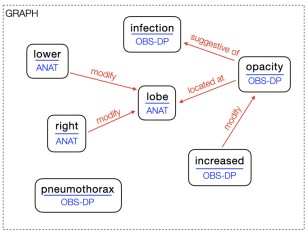}
    \caption{Graph format of the entities and relations derived from Radgraph}
    \label{fig:radg2}
\end{figure}

\section{Data}
Our model utilizes the CheXpert dataset, a large collection of radiology reports, consisting of 220,763 \textbf{MIMIC-CXR} reports. This dataset has been instrumental in advancing research in medical image analysis, particularly through its integration with RadGraph. RadGraph facilitates the extraction and analysis of over 6 million \textit{entities} and 4 million \textit{relationships}, which helps in accurately annotating and parsing radiology reports. These reports have been used in numerous studies to enhance diagnostic algorithms, focusing on diseases visible in chest X-rays and other imaging modalities. The scale of the dataset and the power of RadGraph for annotation parsing make this a highly valuable resource for training machine learning models. By using CheXpert and focusing on 12,716 chest X-ray images \cite{RadGraphCheXpert}, we refined the collection by removing uncertain scans. This left us with 12,549 images for analysis. This subset allows for a more focused examination of radiological findings, ensuring that the data used for machine learning models is of high quality. \par
The prevalence of specific diseases was assessed across the 12,716 chest X-ray images, and the results were broken down into the following findings:

\begin{table}[h!]
\centering
\caption{Prevalence of Diseases in Analyzed Chest X-ray Images}
\setlength{\tabcolsep}{16pt} 
\renewcommand{\arraystretch}{1.5} 
\begin{small} 
\begin{tabular}{|l|c|c|}
\hline
\textbf{Disease} & \textbf{Present} & \textbf{Absent} \\
\hline
Pleural Effusion & 1,289 & 11,260 \\
Pneumothorax & 964 & 11,585 \\
Pneumonia & 971 & 11,578 \\
No Finding & 2,553 & 9,996 \\
\hline
\end{tabular}
\end{small}
\label{tab:prevalence}
\end{table}

These statistics give insight into the distribution of disease occurrences within the analyzed chest X-ray images. It highlights the relative rarity of conditions like \textit{pleural effusion}, \textit{pneumothorax}, and \textit{pneumonia}, as opposed to the more common \textit{no finding} category \cite{Irvin2019}.
By analyzing a subset of 12,549 images (after removing uncertain scans), our model can more reliably detect the diseases presented in the \textit{prevalence report}. The frequency of diseases in the dataset informs us of the imbalance between rare and more frequent conditions, a crucial aspect when developing a multi-label classification model. For instance, \textit{pleural effusion}, \textit{pneumothorax}, and \textit{pneumonia} have relatively low occurrences compared to \textit{no finding}, indicating that models must be able to detect rarer diseases despite their lower prevalence in the data.

These disease prevalence figures will inform the development of the multi-label lung disease classification model, guiding the choice of features, addressing class imbalances, and enhancing model performance.

\section{PreProcessing}
\subsection{Preprocessing with SpaCy for Radgraph}
Natural Language Processing (NLP) plays a critical role in evaluating unstructured medical data, such as radiology reports, with structured annotations. Among the various NLP tools available, machine learning models like spaCy has gained prominence for its speed, flexibility, and pre-trained models capable of handling domain-specific tasks \cite{Eyre2021} through extensions like medspaCy.

\subsection*{Tokenization}
Tokenization is the process of splitting text into smaller units (tokens). SpaCy uses a finite-state automaton to identify tokens:
\[
T = \{t_1, t_2, \dots, t_n\},
\]
where \( T \) represents the sequence of tokens, and \( n \) is the number of tokens. Each character \( c_i \) in the text is evaluated against rules (e.g., punctuation or whitespace) to determine token boundaries \cite{Honnibal2020}. With spaCy's tokenizer, complex language constructs were assessed, including abbreviations, numerical values, and hyphenated terms. In the radiology reports from the CheXpert dataset, tokenization ensured that terms like \textit{“ground-glass opacities”} or \textit{“1.2 cm nodule”} are parsed correctly. The resulting tokens are used as input for subsequent analyses, including parsing and Named Entity Recognition.

\subsection*{Dependency Parsing}
Dependency parsing enables the extraction of syntactic relationships between tokens, which can be used to understand how observations are linked to anatomical structures in radiology reports. By mapping grammatical roles, structured annotations are facilitated, aligning with RadGraph’s requirement to connect findings to specific anatomical regions \cite{Honnibal2020}. This is accomplished when dependency parsing creates a directed graph \( G = (V, E) \), where:
\begin{itemize}
    \item \( V \) represents the tokens (vertices).
    \item \( E \) represents the syntactic dependencies (edges).
\end{itemize}

SpaCy's parser predicts the most probable sequence of transitions \( T \) for a given sentence \( S \):
\[
\text{argmax}_T \; P(T \mid S).
\]
In the sentence of a radiology report \textit{“The left lung shows opacity”}, spaCy identifies the verb \textit{“shows”} as the root and links it to its subject \textit{“left lung”} and object \textit{“opacity”}. By leveraging dependency parsing, we can improve the interpretability of extracted information, which directly supports multi-label classification tasks.

\subsection*{Named Entity Recognition (NER)}
NER identifies spans of text (entities) and classifies them into predefined categories. SpaCy uses models such as BiLSTM-CRF for NER. The objective is to maximize the conditional probability \cite{Honnibal2020}: 
\[
P(Y \mid X) \propto \exp \left( \sum_i e(y_i, x_i) + s(y_i, y_{i-1}) \right),
\]
where:
\begin{itemize}
    \item \( X = \{x_1, x_2, \dots, x_n\} \) represents tokens.
    \item \( Y = \{y_1, y_2, \dots, y_n\} \) represents entity labels.
    \item \( e(y_i, x_i) \) and \( s(y_i, y_{i-1}) \) are emission and transition scores, respectively.
\end{itemize}

\subsection*{Feature Extraction}
Parsed text annotations are represented as feature vectors \( X_{\text{text}} \), where each token \( x_i \in \mathbb{R}^d \) is embedded into a \( d \)-dimensional space. These features are combined with image features \( X_{\text{image}} \) derived from convolutional neural networks (CNNs) \cite{huang2016densely} to form a joint representation:
\[
X = [X_{\text{text}}, X_{\text{image}}].
\]

\subsection*{Multi-Label Classification}
The objective is to predict a set of binary labels \( Y = \{y_1, y_2, \dots, y_m\} \), where \( y_i \in \{0, 1\} \). The classifier maps the combined features \( X \) to probabilities:
\[
P(Y \mid X) = \sigma(WX + b),
\]
where:
\begin{itemize}
    \item \( W \) and \( b \) are the weight matrix and bias vector.
    \item \( \sigma \) is the sigmoid activation function.
\end{itemize}
The loss is computed using binary cross-entropy:
\[
\mathcal{L} = -\frac{1}{m} \sum_{i=1}^m \left[ y_i \log(\hat{y}_i) + (1 - y_i) \log(1 - \hat{y}_i) \right],
\]
where \( \hat{y}_i \) is the predicted probability for label \( i \).

\subsection*{RadGraph Annotation Parsing}
SpaCy extracts structured representations from RadGraph, such as relationships between findings (e.g., "mass") and anatomical sites (e.g., "lung"). These structured outputs improve annotation consistency, directly contributing to enhanced model performance \cite{Dima2022}. To ensure consistency in RadGraph annotations, linguistic similarity between radiology reports was evaluated using noun phrases, verbs, and named entities. This analysis helps identify uniformity in descriptive language across reports, which is essential for accurate and standardized annotations. The Jaccard similarity coefficient was used to calculate similarity scores:
\begin{equation}
    \text{Similarity} = \frac{|A \cap B|}{\max(|A|, |B|)},
\end{equation}
where $A$ and $B$ represent feature sets extracted from two reports. High similarity scores suggest greater overlap in linguistic patterns, aiding in the alignment of annotations with RadGraph's schema. For instance, the similarity between noun phrases highlights whether findings, such as \textit{“opacity in the right lung”}, are described uniformly across different cases.

\subsection*{Frequency Analysis}
\begin{figure}[h]
    \centering
    \includegraphics[width=0.8\linewidth]{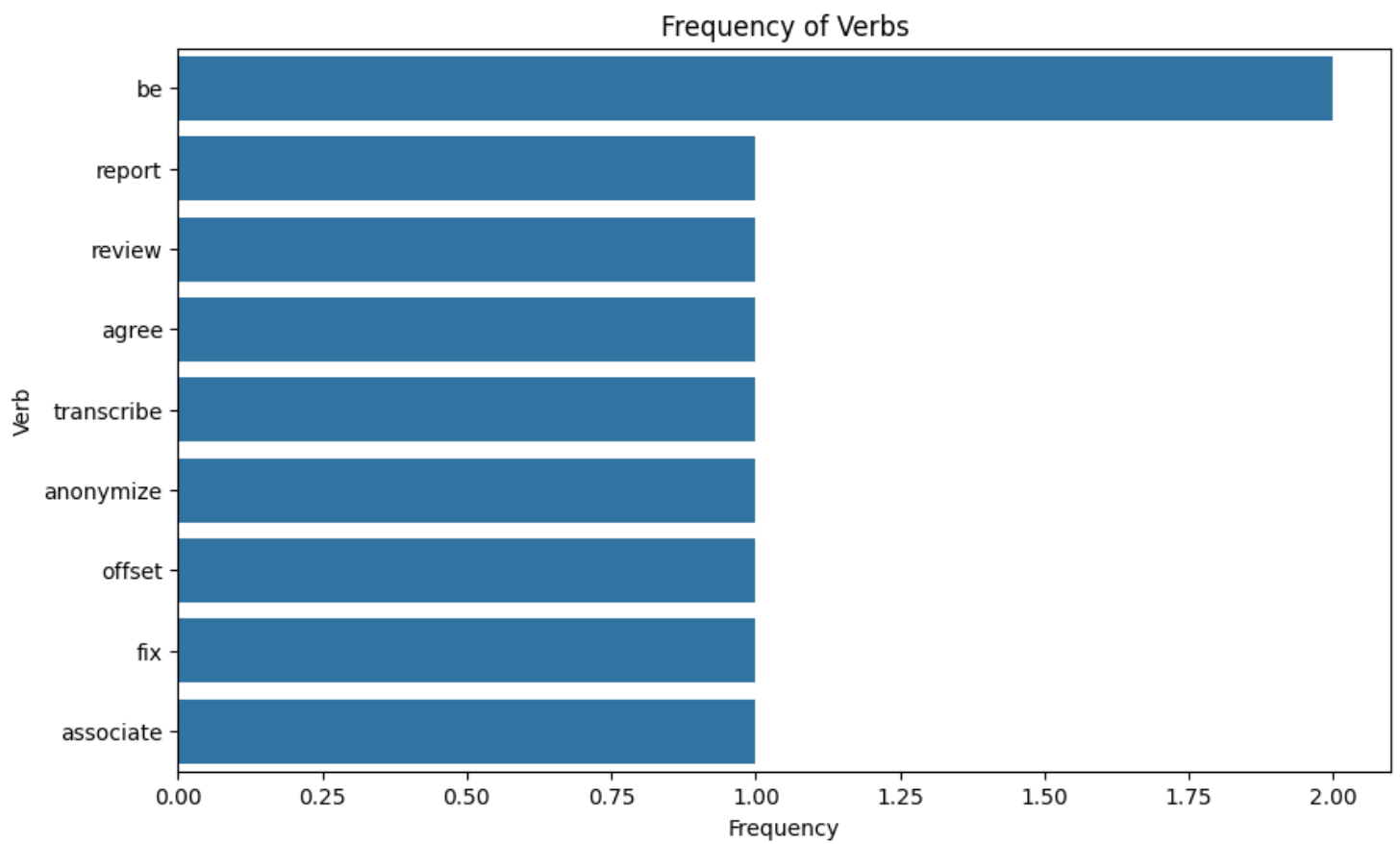}
    \caption{Frequency distribution of verbs across radiology reports.}
    \label{fig:verb_frequency}
\end{figure}

\begin{figure}[h]
    \centering
    \includegraphics[width=0.8\linewidth]{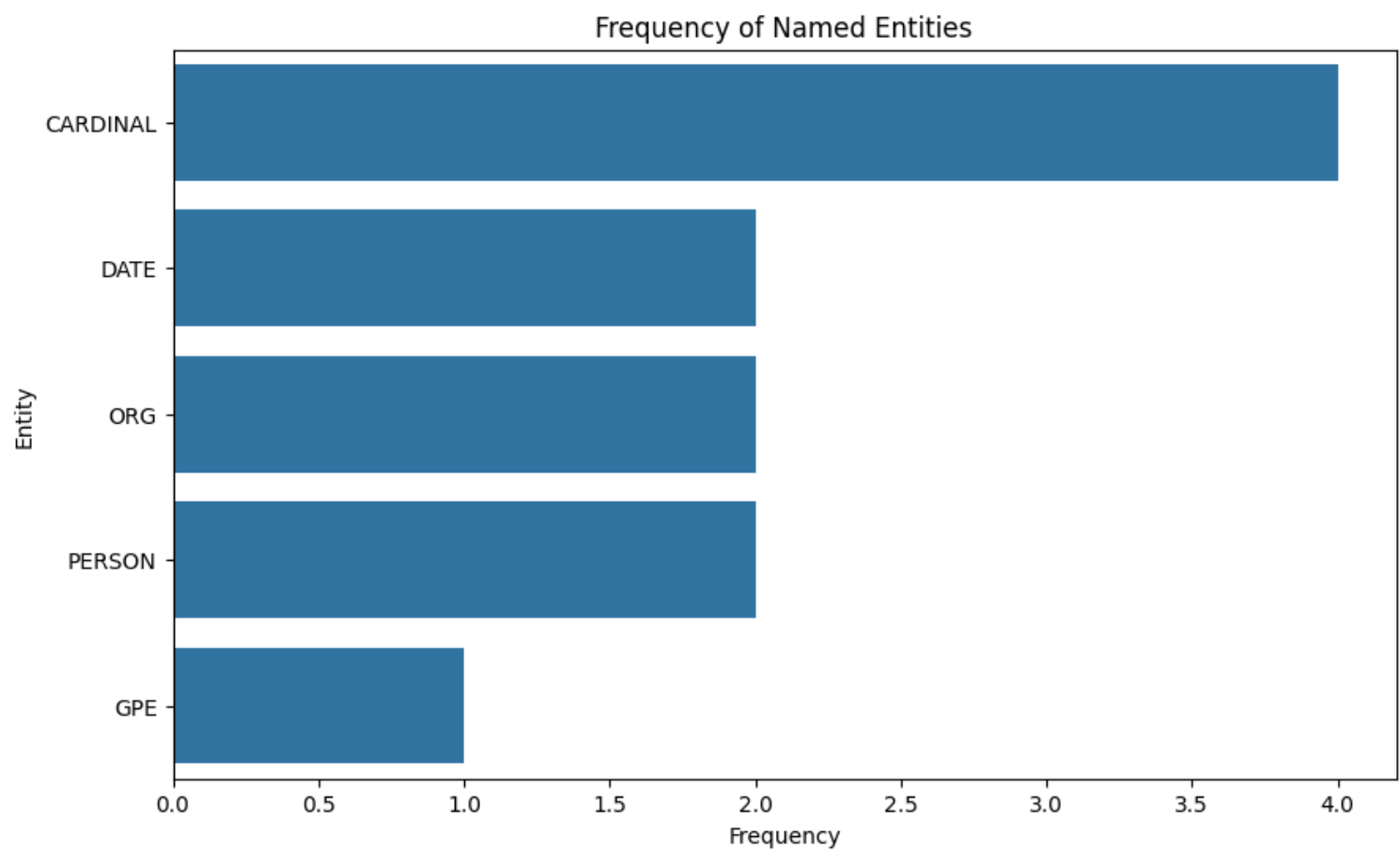}
    \caption{Frequency distribution of named entities across radiology reports.}
    \label{fig:entity_frequency}
\end{figure}

Frequency analysis of verbs and named entities provides insights into the linguistic patterns present in radiology reports. By identifying the most common verbs and entities, we can better understand the vocabulary used to describe diseases and anatomical structures. This information supports the creation of machine learning classifiers for RadGraph annotation tasks \cite{Dima2022}. Figures~\ref{fig:verb_frequency} and~\ref{fig:entity_frequency} illustrate the distributions of verbs and entity labels, which help prioritize terms critical for parsing and annotation.

\subsection{Radgraph Implementation for NLP}
Once spaCy identifies the entities in the text, RadGraph takes over by processing the relationships between these entities. With RadGraph, we analyzed radiology reports by creating relationships between specific entities mentioned within the reports, such as anatomical terms and clinical observations. These relationships are represented in a graph format, allowing for clear visualization of how different entities interact with each other. RadGraph is particularly useful in extracting meaningful insights from unstructured data in medical reports.
\begin{quote}
    Chest 1 view, 8/21/2011:\\
    History: 50 years male, eval pleural effusion reaccum. with clamped chest tube.\\
    Comparison: none.\\
    Impression: 1. Increased right lower lobe opacity, concerning for infection. 2. No evidence of pneumothorax.
\end{quote}

RadGraph processes this report by extracting specific entities such as \textit{right}, \textit{lower}, \textit{lobe}, \textit{opacity}, and \textit{infection}, as well as their relationships, such as \textit{located at} and \textit{suggestive of}. The relationships are visualized as a graph, making it easier to interpret the report.
RadGraph models the entities and relationships in a report as a directed graph \( G = (V, E) \), where:
\[
V = \{v_1, v_2, \dots, v_n\} \quad \text{and} \quad E = \{r_1, r_2, \dots, r_m\}
\]
\( V \) is the set of nodes representing the entities \cite{Eyre2021} (e.g., anatomical terms like "right lung," observations like "opacity") and \( E \) is the set of directed edges representing the relationships between the entities.

\subsection*{Nodes (Entities)}

Each node \( v_i \in V \) represents an entity identified in the text. The types of entities could include \cite{Dima2022}:
\begin{itemize}
    \item \textbf{Anatomy (ANAT)}: Describes body parts (e.g., "right lung," "lower lobe").
    \item \textbf{Observation (OBS)}: Represents clinical observations (e.g., "opacity," "infection").
    \item \textbf{Qualifier (MOD)}: Describes modifications or conditions (e.g., "increased," "concerning").
    \item \textbf{Negation (NEG)}: Denotes absence (e.g., "no evidence").
\end{itemize} 
\vspace{0.1cm}
 Thus, we can model this as:
\[
V = \{v_1, v_2, \dots, v_n\}, \quad v_i \in \{ANAT, OBS, MOD, NEG\}
\]

\subsection*{Edges (Relationships)}

The edges \( E \) represent relationships between the nodes, which are directed and may include types like:
\vspace{0.1cm}
\begin{itemize}
    \item \textbf{Located At}: E.g., "opacity located at the right lung."
    \item \textbf{Suggestive of}: E.g., "opacity suggestive of infection."
    \item \textbf{Modification}: E.g., "increased opacity."
    \item \textbf{Negation}: E.g., "no evidence of pneumothorax."
\end{itemize}
\vspace{0.1cm}
A relationship \( r_k \) between two entities \( v_i \) and \( v_j \) can be mathematically expressed as:
\[
r_k: (v_i \longrightarrow v_j)
\]
\[
r_1: (\text{opacity} \longrightarrow \text{right lung})
\]
Where \( r_k \) represents the type of relationship. If "opacity" (OBS) is located at "right lung" (ANAT) \cite{Lample2016}, we could have:

\subsection*{Graph Construction}

The graph construction involves two major steps:
\begin{enumerate}
    \item \textbf{Entity Extraction}: Identify and classify entities (anatomical terms, observations, etc.).
    \item \textbf{Relationship Extraction}: Extract relationships between the entities based on their co-occurrence in the text and syntactic/semantic patterns.
\end{enumerate}

\subsection*{RadGraph Machine Learning Models}

RadGraph uses several machine learning models in a pipeline to automate the process of extracting entities and relationships and our RadGraph data processing pipeline can be divided into the following main stages:
\begin{itemize}
    \item \textbf{Input Data Processing}: The raw reports are passed into the system.
    \item \textbf{Entity Extraction (using RadGraph)}: The model identifies medical entities in the reports, such as anatomical terms and observations.
    \item \textbf{Condition Labeling}: The extracted entities are then processed, labeled, and categorized based on predefined conditions such as "effusion," "pneumothorax," and "pneumonia."
    \item \textbf{Label Consolidation}: For each report, the labels are consolidated into a structured format after spaCy identifies entities through its named entity recognition.
\end{itemize}

After extracting entities and modeling relationships, the next task is to assign labels to the entities. These labels reflect whether a given condition is \textit{present}, \textit{absent}, or \textit{uncertain} in the report.

Let \( L \) represent the set of possible labels:
\[
L = \{\text{definitely present}, \text{definitely absent}, \text{uncertain}\}
\]
The label assignment function \( f_{\text{label}} \) maps extracted entities \( e_i \) to their corresponding labels \( \ell_i \):
\[
f_{\text{label}}(e_i) = \ell_i \quad \text{where} \quad \ell_i \in L
\]

The label assignment follows predefined rules based on the extracted entities and their context. For example, if the term "Pneumothorax" is marked as "definitely present," it will be assigned the label \( \ell_i = 1 \) (present), while if it is absent, it will be assigned \( \ell_i = -1 \) (absent).

\subsection*{Multi-Report Processing}

RadGraph is designed to process multiple reports. Given a dataset of reports \( \{R_1, R_2, ..., R_m\} \), the system processes each report individually and assigns labels for the medical conditions across all reports \cite{Ngiam2011}. The output is a set of labeled data for the entire dataset.

Let \( f_{\text{multi}} \) represent the function that processes multiple reports:
\[
f_{\text{multi}}(\{R_1, R_2, ..., R_m\}) = \{L_1, L_2, ..., L_m\}
\]
where each \( L_i \) is the output of the label assignment for report \( R_i \).

\subsection{Image Data Preprocessing}
Effective image data preprocessing is vital for preparing datasets like CheXpert for deep learning models, ensuring both the integrity and consistency of input data. In our approach, preprocessing handles various tasks, including image loading, normalization, resizing, and organizing the dataset into a structured format suitable for model training. Each step is designed to maintain high data quality, ensure compatibility with the model, and address potential challenges such as missing or incorrectly formatted images.\par
The CheXpert dataset consists of over 12,716 chest X-ray images, each associated with diagnostic labels such as \textit{Pneumothorax}, \textit{Pneumonia}, \textit{Effusion}, and \textit{No Finding}. The preprocessing pipeline begins by extracting these labels from a CSV file, which links each image to its respective diagnoses. Importantly, the \textit{No Finding} label is derived based on the absence of the conditions represented by other labels, marking images with no detected abnormalities. This derived label facilitates binary classification for cases with no pathological findings.

In addition to labeling, the dataset is structured to focus on frontal scan images only \cite{shih2020chexpert}, with a strict exclusion of studies lacking frontal scans. This step ensures that only relevant data, in which the conditions of interest are most likely to be visible, are included in the training process. By narrowing the dataset in this way, the model can focus on the most informative images \cite{he2016deep}, improving accuracy and generalizability.
\subsection*{Image Resolution and Normalization}

To ensure consistency in input data, each image is resized to a fixed dimension, specifically 320x320 pixels. This resizing process is essential for batch processing in deep learning models \cite{imagenet_cvpr2015}, ensuring that all images are of the same size and meet the requirements of the neural network architecture \cite{vgg16}. Resizing also helps standardize the input dimensions across the dataset, allowing for more efficient training and minimizing computational inconsistencies \cite{zhou2016learning}.
\begin{table}[ht]
\centering
\scriptsize 
\caption{Binary Encoding of Medical Conditions in the Chest X-Ray Dataset}
\setlength{\extrarowheight}{3pt} 
\begin{tabular}{|c|c|c|c|c|}
\hline
\setlength{\tabcolsep}{8pt} 
\renewcommand{\arraystretch}{1.5} 
\textbf{NO FINDING} & \textbf{PNEUMONIA} & \textbf{PNEUMOTHORAX} & \textbf{EFFUSION} \\ \hline
1                   & 0                  & 0                    & 0                \\ \hline
0                   & 1                  & 0                    & 0                \\ \hline
0                   & 0                  & 1                    & 0                \\ \hline
0                   & 0                  & 0                    & 1                \\ \hline
0                   & 0                  & 0                    & 0                \\ \hline
\end{tabular}
\vspace{0.1cm}
\end{table}
Alongside resizing, image pixel values are normalized to fall within a 0–1 range, as opposed to the original 0–255 range. The initial 0-255 range  is typical found in images that are not normalized. The process of normalization involves scaling these pixel values to a range between 0 and 1 (or another range), which can improve the performance and stability of deep learning models by ensuring consistent input data values \cite{imagenet_cvpr2015}. This normalization process scales the pixel values, making them more suitable for neural network models \cite{he2017imbalanced}, as most deep learning frameworks perform better when inputs are scaled within a smaller range. The normalization step enhances the model's ability to converge during training, speeding up the learning process and improving performance.\par
In this preprocessing step, only frontal chest X-ray scans were considered for analysis, ensuring consistent imaging perspectives across the dataset. Studies lacking frontal scans were excluded to maintain the quality and relevance of the dataset. The binary encoding of conditions, represented in Table 2, facilitates the multi-label classification approach used in this study.

\section{Dataset Imbalance and Mitigation Strategies}

One of the most significant challenges encountered during this study was the class imbalance within the dataset. The "No Finding" class was overwhelmingly represented compared to other diagnostic categories such as Pneumonia, Pneumothorax, and Effusion. This imbalance led to a biased model that predominantly predicted "No Finding," as doing so maximized accuracy under these skewed conditions.

To address this, several strategies were explored. Initially, \textit{samplers} were used to ensure equal representation of each class during training by dynamically resampling the dataset. Additionally, a \textit{weighted loss function} was implemented to penalize incorrect predictions more heavily for underrepresented classes, incentivizing the model to learn these minority classes more effectively. Despite these efforts, neither approach sufficiently improved the model's performance.

Given the constraints of time and resources, a more drastic solution was adopted. A significant portion (91\%) of the "No Finding" data was dropped, alongside 30\% of the Pleural Effusion data. This reduction aimed to bring the class distributions closer to balance while maintaining a sufficient quantity of diverse examples for training. Although not an ideal solution, this approach allowed the model to better generalize across all classes and avoid the over-representation bias inherent in the original dataset.
To address the class imbalance, a weighted cross-entropy loss function was utilized:
\[
L = -\sum_{i=1}^{N} w_i \cdot y_i \log(\hat{y}_i)
\]
where:
\begin{itemize}
    \item \(N\) is the total number of samples,
    \item \(w_i\) is the weight assigned to class \(i\),
    \item \(y_i\) is the true label, and
    \item \(\hat{y}_i\) is the predicted probability.
\end{itemize}
\vspace{0.1cm}\par
However, this approach did not yield satisfactory results, requiring a different strategy. (91\%) of the \textit{No Finding} data was dropped, alongside 30\% of the \textit{Pleural Effusion} data. This reduction aimed to bring the class distributions closer to balance while maintaining a sufficient quantity of diverse examples for training. This approach allowed the model to better generalize across all classes and avoid the over-representation bias inherent in the original dataset.

\begin{figure}[h!]
    \centering
    \includegraphics[width=9cm]{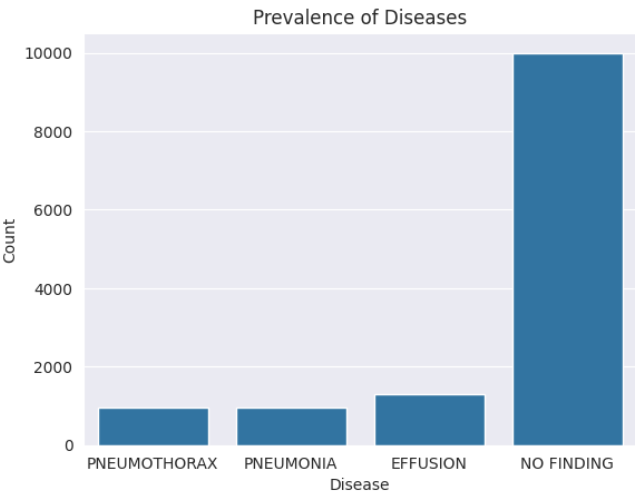}
    \caption{Class distribution before dataset reduction.}
    \label{fig:dataset_reduction}
\end{figure}

\begin{figure}[h!]
    \centering
    \includegraphics[width=9cm]{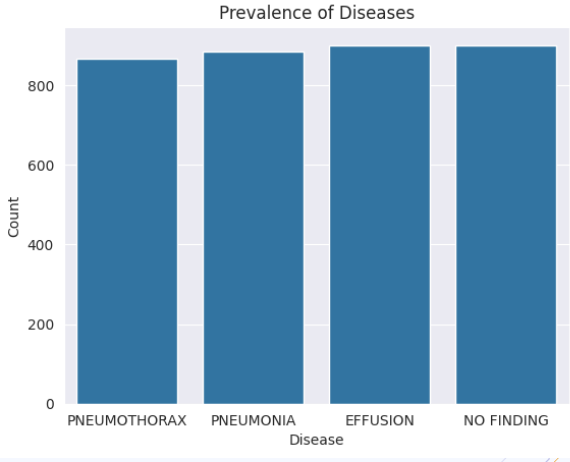}
    \caption{Class distribution before dataset reduction.}
    \label{fig:dataset_reduction}
\end{figure}

\section{Model Architecture}

The proposed model architecture implements a classification pipeline to diagnose pneumonia from chest X-ray images. The architecture integrates three pretrained models—\textbf{VGG16}, \textbf{ResNet50}, and \textbf{ConvNeXt Large}—to extract high-level features, followed by a custom classifier to predict the presence of pneumonia. This section details the components and operations of the model.

The input to the model is a chest X-ray image \cite{simonyan2014very}, mathematically represented as:
\[
\mathbf{X} \in \mathbb{R}^{H \times W \times C},
\]
where:
\begin{itemize}
    \item \( H \): Height of the image,
    \item \( W \): Width of the image,
    \item \( C \): Number of channels (typically \( C = 1 \) for grayscale X-ray images).
\end{itemize}
\vspace{0.1cm}\par
This section discusses the input after undergoing preprocessing steps, such as resizing and normalization, before being passed to the following pretrained models. Three feature extraction backbones are employed:
\begin{enumerate}
    \item \textbf{VGG16 Base}: A convolutional neural network (CNN) \cite{simonyan2014very} with 16 layers, characterized by its uniform architecture and stacked convolutional blocks \cite{szegedy2015going}.
    \item \textbf{ResNet50 Base}: A deeper CNN that utilizes residual connections to mitigate the vanishing gradient problem \cite{he2016deep}. Residual connections are defined as:
    \[
    \mathbf{y} = F(\mathbf{x}, \{\mathbf{W}_i\}) + \mathbf{x},
    \]
    where \( F \) is the residual mapping, \( \mathbf{W}_i \) are learnable weights, and \( \mathbf{x} \) is the input to the residual block.
    \item \textbf{ConvNeXt Large Base}: A modern CNN architecture inspired by transformer \cite{liu2022convnet} design principles \cite{he2016deep}, optimized for hierarchical feature extraction.
\end{enumerate}
\vspace{0.1cm}
Each pretrained model processes the input \( \mathbf{X} \) to produce a feature vector:
\[
\mathbf{f}_i = g_i(\mathbf{X}),
\]
where \( g_i \) represents the feature extraction function for the \( i \)-th model.

The extracted feature vectors \( \mathbf{f}_1, \mathbf{f}_2, \mathbf{f}_3 \) are combined to form a unified feature representation \( \mathbf{f}_{\text{combined}} \), which is passed to a custom classifier. This classifier maps the features to a binary output that predicts the presence (\( y = 1 \)) or absence (\( y = 0 \)) of pneumonia. 
To address class imbalance in the dataset, a weighted binary cross-entropy loss was employed:
\[
\mathcal{L} = -\frac{1}{N} \sum_{i=1}^{N} w_{y_i} \left[ y_i \log \hat{y}_i + (1 - y_i) \log (1 - \hat{y}_i) \right],
\]
where:
\begin{itemize}
    \item \( N \): Number of samples in the batch,
    \item \( y_i \): True label (\( 0 \) or \( 1 \)),
    \item \( \hat{y}_i \): Predicted probability,
    \item \( w_{y_i} \): Class weight, emphasizing the minority class.
\end{itemize}
\begin{figure}[h!]
    \centering
    \includegraphics[width=8cm]{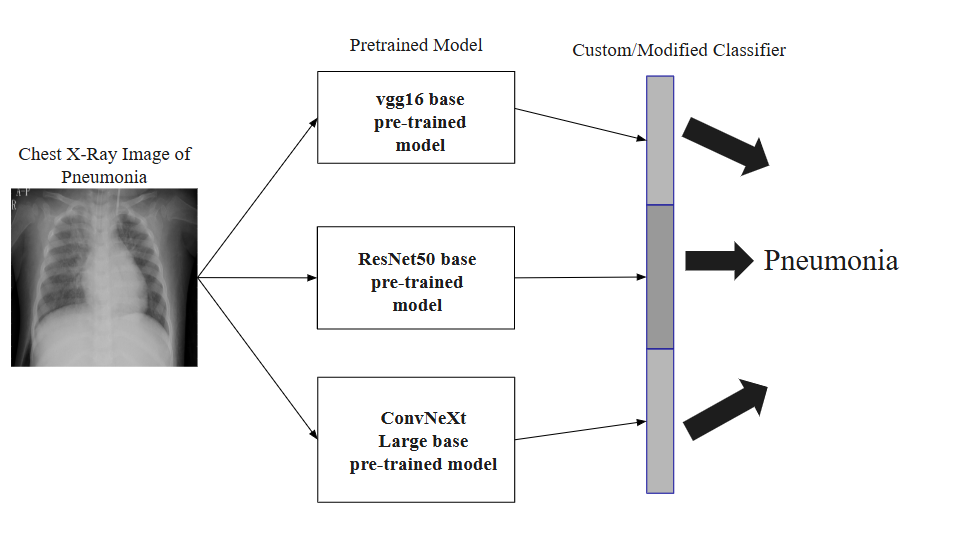}
    \caption{Model architecture}
    \label{fig:dataset_reduction}
\end{figure}
\vspace{0.5cm}
The architecture operates as follows:
\begin{enumerate}
    \item \textbf{Input Preprocessing}: Resize and normalize the chest X-ray images.
    \item \textbf{Feature Extraction}: Extract features using VGG16, ResNet50, and ConvNeXt Large pretrained models.
    \item \textbf{Classification}: Combine features and predict pneumonia probabilities using the custom classifier.
    \item \textbf{Output}: Binary label indicating the presence or absence of pneumonia.
\end{enumerate}
\subsection*{Validation Metrics across Epochs}
\begin{figure}[h]
    \centering
    \includegraphics[width=0.5\textwidth]{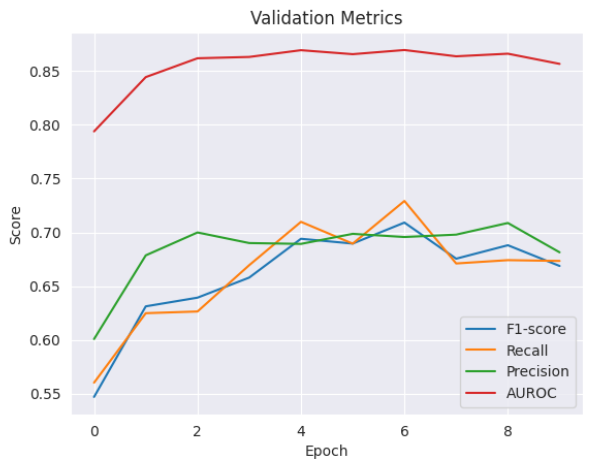}
    \caption{Validation Metrics (AUROC, F1-Score, Precision, Recall) over epochs}
    \label{fig:loss}
\end{figure}

The first graph (Figure~\ref{fig:loss}) presents the trends in training and validation loss over the epochs. The training loss decreases consistently as the epochs progress, indicating that the model is learning effectively and fitting the training data. The validation loss initially decreases, which reflects the model's ability to generalize to unseen data. However, after epoch 6, the validation loss begins to fluctuate, suggesting that further training might lead to overfitting. A red vertical line at epoch 6 marks the optimal stopping point where the validation loss is lowest, demonstrating the application of early stopping to prevent overfitting and preserve the model's generalization capability.\par
\begin{figure}[h]
    \centering
    \includegraphics[width=0.4\textwidth]{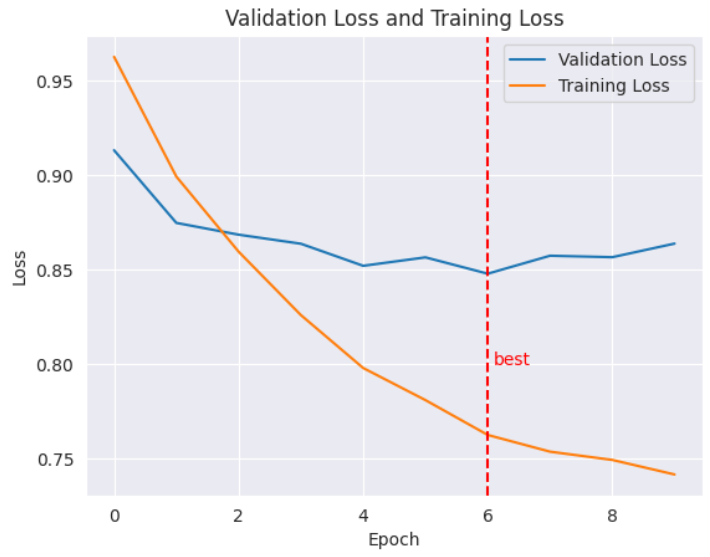}
    \caption{Validation Loss and Training Loss across epochs}
    \label{fig:metrics}
\end{figure}

The second graph (Figure~\ref{fig:metrics}) illustrates the trends in validation metrics, including AUROC, F1-score, Precision, and Recall, across epochs. The AUROC metric remains consistently high throughout training, peaking early and stabilizing around 0.85, which indicates the model's strong ability to discriminate between classes. In contrast, the other metrics show more variability. Recall increases sharply in the initial epochs but fluctuates afterward, reflecting inconsistent sensitivity to positive cases. Similarly, Precision and F1-score exhibit fluctuations across epochs, highlighting the challenge of balancing true positives, false positives, and false negatives in the presence of imbalanced classes.\par
Collectively, these results demonstrate that the model achieves strong discriminative performance, as evidenced by the high AUROC, despite the variability in other metrics. The fluctuations in precision, recall, and F1-score emphasize the challenges associated with class imbalance \cite{goodfellow2016deep}, where improving one metric may degrade another. Early stopping at epoch 6 ensures that the model performs optimally without overfitting. These findings highlight the importance of addressing dataset imbalances and monitoring validation metrics closely to optimize the model's performance.
\begin{table}[h!]
\centering
\caption{Comparison of different methods based on Precision, Recall, F1-Score, and AUROC.}
\renewcommand{\arraystretch}{1.3} 
\setlength{\tabcolsep}{5pt}       
\fontsize{9pt}{11pt}\selectfont  
\begin{tabular}{|c|c|c|c|c|}
\hline
\textbf{Method}     & \textbf{Precision} & \textbf{Recall} & \textbf{F1-Score} & \textbf{AUROC} \\ \hline
ResNet-18           & 0.657              & 0.640           & 0.638             & 0.836          \\ \hline
ResNet-50           & 0.643              & 0.616           & 0.627             & 0.814          \\ \hline
DenseNet            & 0.705              & 0.646           & 0.664             & 0.849          \\ \hline
MobileNetv2         & 0.624              & 0.589           & 0.605             & 0.813          \\ \hline
MobileNetv3         & 0.686              & 0.6319          & 0.641             & 0.848          \\ \hline
\textbf{ConvNeXt}            & \textbf{0.699}              & \textbf{0.689}           & \textbf{0.690}             & \textbf{0.866}          \\ \hline
\end{tabular}
\label{tab:model_comparison}
\end{table}
\subsection{Transfer Learning}
Transfer learning is a technique in deep learning where a model trained on a large \cite{zhuang2019survey}, general-purpose dataset (e.g., ImageNet) \cite{deng2009imagenet} is repurposed for a related but distinct task, such as medical image classification. The approach leverages the previously learned features, reducing the need for large datasets and computational resources for the new task.

In transfer learning, models extract generic features from the lower layers (such as edges and textures in images) and adapt the higher layers to task-specific features. There are two common strategies:

\begin{itemize}
    \item \textbf{Feature Extraction:} Pretrained models serve as fixed feature extractors by freezing all layers and replacing only the final layers.
    \item \textbf{Fine-Tuning:} Some layers are ``unfrozen'' to retrain on the new dataset, adapting the pretrained weights while preserving the foundational features.
\end{itemize}

Mathematically, if the pretrained model optimizes a loss function $L$ using weights $W$, then transfer learning seeks to minimize the task-specific loss:
\[
L_{\text{new}} = f(W_{\text{pretrained}}) + \lambda \cdot \text{Regularization}
\]
where $f(W_{\text{pretrained}})$ represents fine-tuning of pretrained weights for the new data.

PyTorch provides extensive support for pretrained models via the torchvision library~\cite{paszke2019pytorch}~\cite{marcel2010torchvision}~\cite{albardi2021torchvision}. The pretrained weights are generated by training models on millions of diverse images to learn representations of basic visual patterns. The workflow involves:
\begin{itemize}
    \item \textbf{Layer Freezing:} Using model.parameters() in PyTorch, layers can be frozen to preserve pretrained weights.
    \item \textbf{Custom Output Layers:} The final fully connected (or softmax) layers are often replaced to fit the target problem's number of classes.
    \item \textbf{Learning Rate Adjustment:} A smaller learning rate is used during fine-tuning to prevent overwriting the pretrained weights too rapidly.
\end{itemize}

The pre-trained weights were derived from extensive training processes that included optimized learning schedules and data augmentation techniques. PyTorch uses the ImageNet-1k classification task to ensure the models achieve high generalization performance across diverse visual categories. By training on ImageNet, these models capture hierarchical feature representations, such as edges, textures, and shapes, in earlier layers, which are crucial for transfer learning to more specific applications, such as predicting chest diseases from X-rays. This approach significantly reduces the computational cost and data requirements compared to training models from scratch, while maintaining state-of-the-art performance in various tasks \cite{zhuang2019survey} \cite{deng2009imagenet}.

\section{Model Results}
\subsection{AUROC curve}
\begin{figure}[h]
    \centering
    \includegraphics[width=0.4\textwidth]{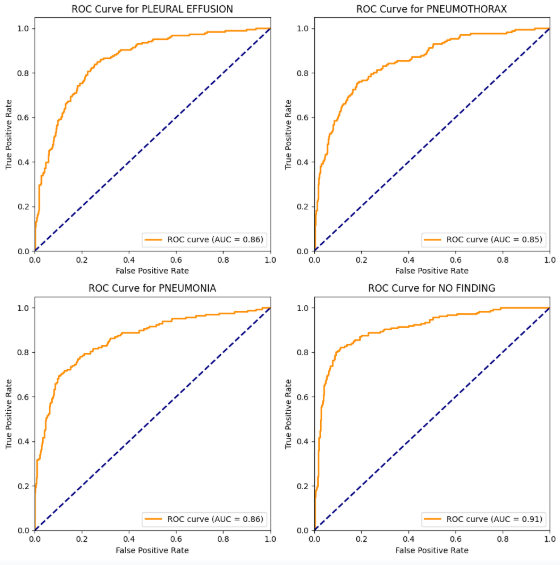}
    \caption{Validation Metrics (AUROC, F1-Score, Precision, Recall) over epochs.}
    \label{fig:result}
\end{figure}

Figure~\ref{fig:result} displays Receiver Operating Characteristic (ROC) curves for four medical diagnostic categories: \textit{Pleural Effusion}, \textit{Pneumothorax}, \textit{Pneumonia}, and \textit{No Finding}. Each plot includes the corresponding Area Under the Curve (AUC) value, which provides a quantitative measure of the model's ability to distinguish between positive and negative cases for each condition. \par
The ROC curves are well above the diagonal line of no-discrimination (blue dashed line), which indicates that the model performs significantly better than random guessing for all four conditions. The AUC values, ranging from 0.85 to 0.91, suggest that the model is highly effective across these categories \cite{fawcett2006introduction}. Higher AUC values indicate stronger diagnostic performance, with \textit{No Finding} achieving the highest AUC of 0.91, suggesting that the model excels at identifying cases where no abnormalities are present.
\begin{table}[h!]
\centering
\caption{Specific performance insights based on ROC curves and AUC values for each condition.}
\renewcommand{\arraystretch}{1.5} 
\setlength{\tabcolsep}{10pt} 
\begin{tabular}{|>{\raggedright\arraybackslash}p{3.5cm}|>{\raggedright\arraybackslash}p{3cm}|}
\hline
\textbf{Condition}      & \textbf{AUROC values}                                                                                      \\ \hline
Pleural Effusion        & 0.86  \\ \hline
Pneumothorax            & 0.85 \\ \hline
Pneumonia               & 0.86 \\ \hline
No Finding              & 0.91 \\ \hline
\end{tabular}
\label{tab:performance-insights}
\end{table}
\subsection{Confusion Matrix}
Each confusion matrix quantifies the model's predictions by comparing the true labels (rows) with the predicted labels (columns), allowing us to understanding of the model's accuracy, sensitivity, and misclassification trends across different conditions. \par
For the \textbf{Pleural Effusion} category, the model exhibits the \textit{highest number of false positives}, with 87 cases incorrectly classified as pleural effusion when the condition was not present. This suggests that the model is overly sensitive to features indicative of pleural effusion, leading to frequent misclassification. Additionally, the model has a relatively high number of false negatives (46 cases), where pleural effusion was missed. These results indicate that pleural effusion is one of the more challenging categories for the model to classify, potentially due to overlapping features with other conditions. This leads to a significant trade-off between sensitivity and precision, requiring refinement to reduce false positives while maintaining detection capability.

In contrast, the \textbf{Pneumothorax} category has the \textit{fewest false negatives}, with only 47 cases missed, suggesting that the model is relatively effective at detecting this condition. However, it still struggles with false positives, with 74 cases incorrectly classified as pneumothorax. This indicates that while the model demonstrates strong sensitivity for pneumothorax, it requires improvement in specificity to avoid over-diagnosis. Addressing these false positives could involve refining the features used for pneumothorax detection or adjusting classification thresholds.

The \textbf{No Finding} category demonstrates the \textit{lowest number of false positives} (38 cases), reflecting high specificity in identifying normal cases. Furthermore, the false negative count (40 cases) is also among the lowest across all categories. This suggests that the model performs particularly well when identifying cases with no abnormal findings, maintaining a balance between sensitivity and precision. The performance of the model in this category may stem from the distinctiveness of normal cases compared to pathological findings.\par
\begin{figure}[h]
    \centering
    \includegraphics[width=0.4\textwidth]{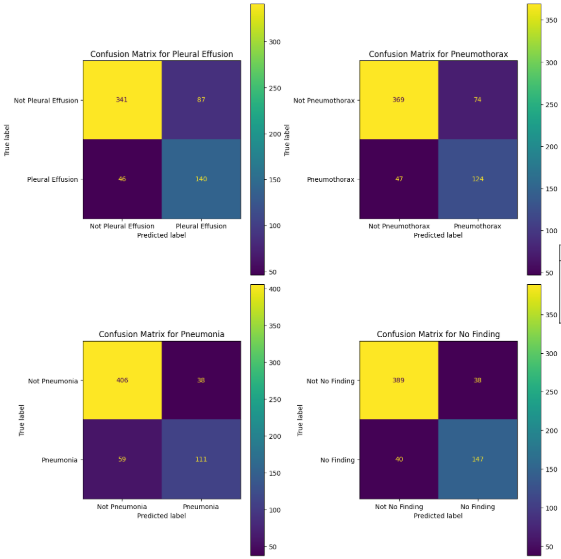}
    \caption{Outputted Confusion Matrix}
    \label{fig:results}
\end{figure}
\begin{table}[h!]
    \centering
    \caption{Confusion matrix metrics for each condition. TP: True Positives, FP: False Positives, TN: True Negatives, FN: False Negatives.}
    
    \setlength{\tabcolsep}{10pt} 
    \renewcommand{\arraystretch}{1.5} 
    {\normalsize 
    \begin{tabular}{|l|c|c|c|c|}
        \hline
        \textbf{Method} & \textbf{TP} & \textbf{FP} & \textbf{TN} & \textbf{FN} \\
        \hline
        Pleural Effusion & 140 & 87 & 341 & 46 \\
        \hline
        Pneumothorax     & 124 & 74 & 369 & 47 \\
        \hline
        Pneumonia        & 111 & 38 & 406 & 59 \\
        \hline
        No Finding       & 147 & 38 & 389 & 40 \\
        \hline
    \end{tabular}}
    \label{tab:metrics}
\end{table}

The \textbf{Pleural Effusion} category has the most false positives, suggesting challenges with over-diagnosis \cite{bhatnagar2014confusion}, while \textbf{Pneumonia} has the highest number of false negatives, indicating difficulties with sensitivity. The \textbf{Pneumothorax} category strikes a balance, with the fewest false negatives but a moderate number of false positives.
\section{Grad Cam: Model evaluation}
\begin{figure}[h]
    \centering
    \includegraphics[width=0.4\textwidth]{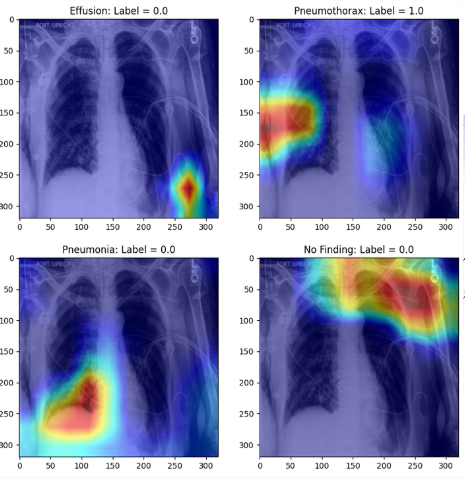}
    \caption{Light heatmap around the collapsed portion with No Findings}
    \label{fig:heat}
\end{figure}

Grad-CAM (Gradient-weighted Class Activation Mapping) was also used for visualization to enhance the deep learning model. By leveraging the gradients flowing into the final convolutional layers, Grad-CAM identifies which regions of an image contribute most to a model's predictions \cite{chattopadhay2018gradcam}. This approach overlays heatmaps on the original image, visually highlighting areas of focus for the model, enabling clinicians to understand the "reasoning" behind its decisions \cite{selvaraju2017grad}.
Upon implementation of Grad-CAM, our visualizations added a layer of interpretability to the model’s outputs. For Pneumothorax, heatmaps clearly focused on collapsed lung regions as seen in Figure~\ref{fig:heat}, providing clinically relevant insights. Similarly, for Pleural Effusion, the model’s attention was directed toward areas of fluid buildup, demonstrating its capacity to localize abnormalities accurately. \par
\begin{figure}[h]
    \centering
    \includegraphics[width=0.4\textwidth]{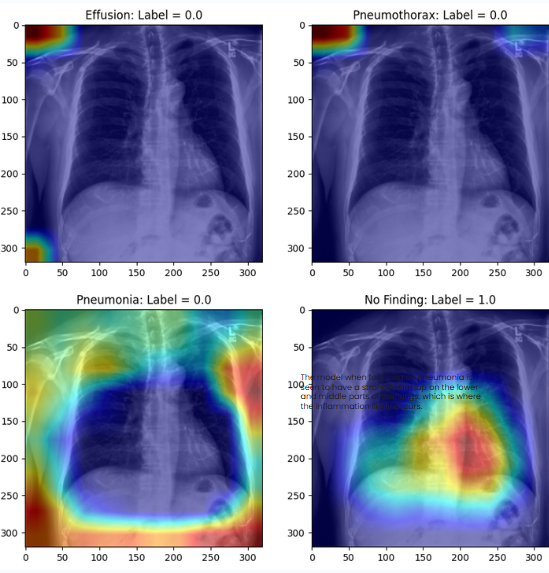}
    \caption{Light heatmap around the collapsed portion with pneumothorax}
    \label{fig:heat2}
\end{figure}

In cases of pneumonia, Grad-CAM highlighted inflamed lower and middle lung regions, while in the "No Finding" scenarios seen in Figure~\ref{fig:heat2}, the attention was more diffuse, sometimes reflecting areas less clinically relevant. These insights underline the importance of combining tools with deep learning models to build clinician trust.
Grad-CAM works by computing a weighted combination of the gradients of a target class with respect to the feature maps of the final convolutional layer. This combination highlights the most important regions in the input image. With our data, Grad-CAM was tasked to perform the following: 

\subsection*{Gradient Calculation}
The gradient of the score \( y^c \) (corresponding to class \( c \)) was calculated with respect to the feature maps \( A^k \) of the final convolutional layer, where \( k \) indexes the feature maps \cite{shrikumar2017learning}:
\[
\frac{\partial y^c}{\partial A^k}
\]

\subsection*{Global Average Pooling}
A global average pooling was then performed on these gradients to obtain the weights \( \alpha_k^c \):
\[
\alpha_k^c = \frac{1}{Z} \sum_{i} \sum_{j} \frac{\partial y^c}{\partial A_{ij}^k}
\]
where \( Z \) is the number of pixels in the feature map (spatial dimensions \cite{lin2013network} \( i, j \)).

\subsection*{Weighted Feature Map Combination}
The weighted combination of the feature maps produces the class-discriminative localization map \cite{selvaraju2017grad}\( L^c_\text{Grad-CAM} \):
\[
L^c_\text{Grad-CAM} = \text{ReLU} \left( \sum_k \alpha_k^c A^k \right)
\]
The ReLU ensures that only features with a positive influence on the target class are included.The resulting localization map \( L^c_\text{Grad-CAM} \) is upsampled to match the size of the input image and overlaid as a heatmap \cite{shrikumar2017learning}. Overall, Grad-CAM contributed to our model and overall product by:\begin{enumerate}
\vspace{0.1cm}
    \item Visualizing the regions contributing to the model's predictions, Grad-CAM validates the alignment of our AI outputs with medical knowledge. This would be especially important in critical domains like healthcare, where clinician trust is paramount.
    
    \item Assisting error analysis by revealing whether incorrect predictions stem from focusing on irrelevant regions or failing to capture relevant features.
    
    \item Providing insights that can inform model refinement. For instance, if heatmaps consistently highlight irrelevant areas, it may indicate a need for better preprocessing, dataset balancing, or architectural changes.
    
    \item Enabling AI models to function as collaborative tools by providing clinicians with visual explanations. This can enhance confidence in automated systems and promote their adoption in practice.
\end{enumerate}
\vspace{0.1cm}
By combining interpretability tools like Grad-CAM with high-performing deep learning models, this study demonstrates a step toward reliable and trustworthy AI systems for medical imaging.
\section{Discussion}
The ConvNeXt model was the best-performing network, showing capabilities for handling both the complexity of medical imaging and the challenges of dataset imbalance. With a validation accuracy of 92

Among the reasons for its success is the robust handling of imbalance in the dataset. Medical datasets often exhibit class imbalance, where pathological cases are underrepresented compared to normal cases. To handle this challenge, ConvNeXt incorporated a weighted cross-entropy loss function wherein higher penalties for misclassifications were levied on samples of the minority class. Besides that, data augmentations during training like random rotations, flipping, and intensity variations helped present the minority class features in a much more comprehensive way.

Major architectural changes were crucial in making the ConvNeXt model perform exceptionally well. Using transformer-based principles but retaining the computational efficiency of convolutional neural networks, ConvNeXt introduced depth-wise convolutions and layer normalization to enhance feature extraction \cite{woo2023convnext}. This hierarchical design allowed the model to capture both local and global features. Besides the architectural novelty, another key advantage of the ConvNeXt model was its scalable nature and robust pretraining on large datasets \cite{woo2023convnext}. This is definitely because, with the pre-trained weights, it learned the general features of images and adapted well to the smaller domain-specific dataset of X-rays. Another essential factor was its compatibility with advanced data augmentation techniques. ConvNeXt proved resilient against input data variances, such as noise or slight distortions, which are often disruptive for less sophisticated models. These attributes combined enabled the performance of ConvNeXt to outperform others in both performance metrics and interpretability.

\begin{figure}[h]
    \centering
    \includegraphics[width=0.4\textwidth]{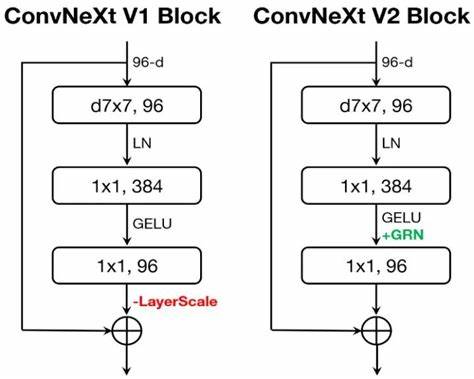}
    \caption{CONVNEXT architecture}
    \label{fig:1}
\end{figure}

GradCAM analysis provided strong support for ConvNeXt as a diagnostic tool by underlining its decision-making process. Heatmaps generated by GradCAM pinpointed vital regions within the X-rays that presented opacities, fractures, or other anomalies in structure and shape that may be indicative of a pathological condition. The congruence of these highlighted regions with the radiologists' annotations underlined the model's reliability. GradCAM presented further opportunities for possible improvements by analyzing misclassified samples \cite{selvaraju2019grad}. Sometimes, it also paid attention to non-critical regions or even noise for its predictions. These observations show the necessity of refining the attention mechanism in this model to make it more interpretable and accurate. Generally, GradCAM demonstrated that ConvNeXt did not only excel in prediction accuracy but was also explainable, which can make it trustworthy in clinical applications \cite{selvaraju2019grad}.

Though ConvNeXt demonstrated excellent performance, several directions can still be followed for its improvement in the future. Experiments on other architectures, like ViT or Swin Transformers, may yield useful comparisons \cite{dosovitskiy2020transformers} \cite{liu2021swin}. These models are renowned for their capability to show long-range dependencies, which may help improve the detection of subtle features in medical imaging. Hybrid approaches, which couple CNNs with transformer-based methods, can also provide an optimal balance between localization and global context awareness. Furthermore, the incorporation of NLP tools provides an interesting line of investigation for raising the accuracy of diagnostics. Models like BioBERT or ClinicalBERT could be applied to radiology reports or metadata to give insights complementary to image-based predictions. Incorporation of patient metadata, like age, medical history, or genetic predispositions, would further contextualize predictions, improving the model's clinical utility \cite{lee2019biobert} \cite{huang2019clinicalbert}. Other techniques, such as SMOTE or GANs, might better address the imbalance in the dataset by generating realistic samples for underrepresented classes \cite{chawla2002smote} \cite{rangwani2021class}.

\begin{figure}[h]
    \centering
    \includegraphics[width=0.4\textwidth]{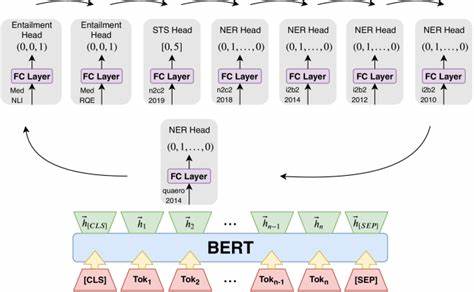}
    \caption{ClinicalBERT architecture}
    \label{fig:1}
\end{figure}

The success of ConvNeXt opens the door to a wide range of future applications in medical diagnostics. The model could be adapted to other radiographic datasets-such as mammograms, chest CTs, or bone density scans-to assess its versatility \cite{woo2023convnext}. Fine-tuning the model on different medical modalities would allow it to tackle diverse diagnostic challenges. Beyond X-rays, ConvNeXt could be extended to analyze other imaging modalities like MRIs or ultrasounds, addressing broader medical needs. Real-time deployment of the model in edge computing environments could revolutionize diagnostic workflows, particularly in low-resource areas. Integrating LIME with GradCAM techniques of explainability could still better capture a fine view of how the model would be working \cite{rangwani2021class}. Allowing radiologists to dynamically interact with model predictions and explanations would therefore allow increased trust in such decision-making processes. Advances stand to not only make diagnostic processing more accurate but also to democratize access to advanced medical capabilities to benefit global health outcomes significantly.

\section{Conclusion}
This study demonstrates the effective deployment of multilabel classification models for diagnosing lung diseases, achieving an F1 score of 0.69 and an AUROC of 0.86. These metrics indicate a strong balance between sensitivity and specificity, emphasizing the potential of deep learning in automating complex medical diagnoses. SpaCy, Radgraph, and other image preprocesing techniques were valuable in preprocessing the data which could be fed into pre-trained models. \par
These NLP tools proved to be effective in helping better evaluate images with our proposed machine learning model. Among the tested architectures, ConvNext and DenseNet models outperformed VGG and other models, highlighting the importance of leveraging modern architectures optimized for high-dimensional image data in medical imaging.
The use of Grad-CAM was an added feature, bringing another layer of analysis and assistance for future use which would enable the identification of clinically relevant features such as collapsed lung regions for Pneumothorax and fluid buildup in Pleural Effusion. However, attention diffusion in "No Finding" cases revealed limitations in handling less distinct patterns, suggesting areas for refinement. These insights underline the value of explainability tools in fostering clinician trust and improving diagnostic accuracy.

Future advancements in this field could integrate multimodal approaches, combining vision models with natural language processing (NLP) tools. Parsing radiology reports, metadata, and patient history into the classification pipeline can contextualize predictions and address potential blind spots in imaging-only models. Such hybrid frameworks could enhance diagnostic performance by providing a more comprehensive understanding of patient conditions.

This study underscores the significance of combining state-of-the-art vision models, interpretability tools like Grad-CAM, and future multimodal approaches to build reliable, clinically applicable AI systems for radiology. By addressing challenges such as the classification of uncertain conditions and enhancing the pipeline with NLP tools, these systems hold the potential to transform medical diagnostics by improving accuracy, efficiency, and trustworthiness, ultimately benefiting both clinicians and patients.

\section{Acknowledgements}
We would like to express our sincere gratitude to all those who contributed to the success of this work. First and foremost, we thank our supervisor, Rhea Malhotra, for her invaluable guidance and support throughout the course of this project. A special acknowledgment is given to Ethan Poon, who primarily authored this paper. We would also like to thank the Stanford Center for AI in Medicine and Imaging (AIMI) for the opportunity to intern and perform this study. The expertise and insightful feedback of our advisor and university professors were instrumental in shaping the direction of our research. We also wish to acknowledge the contributions of Stanford University for providing the necessary resources and support for this study. The integration of NLP and imagine techniques proves to be a key area of study, revolutionizing the way conditions are identified. Finally, we would like to thank our families for their support and patience during this study.
\bibliographystyle{IEEEtran}
\bibliography{references} 
\nocite{*}

\end{document}